\documentclass[letterpaper, 10 pt, conference]{ieeeconf}    

\IEEEoverridecommandlockouts                              

\usepackage{graphicx}
\usepackage{lineno}
\usepackage[hidelinks]{hyperref}
\hypersetup{hidelinks}
\usepackage{xcolor}
\usepackage{pgfplots}
\pgfplotsset{compat=1.18} 
\usepackage{tikz}
\usetikzlibrary{positioning}
\usetikzlibrary{shapes.geometric}
\usetikzlibrary{shapes.misc}
\usetikzlibrary{external}
\usepackage{paralist}
\usepackage{mathbbol}
\usepackage{amssymb}    
\usepackage{bm}
\usepackage{xcolor}
\usepackage{siunitx}
\usepackage{multirow}
\usepackage{epstopdf}
\usepackage{todonotes}
\usepackage{epstopdf}
\usepackage{epsfig}
\usepackage{float}
\usepackage{amsmath} 
\usepackage{amssymb}  
\usepackage{mathtools}
\usepackage[acronym,nomain,nonumberlist]{glossaries}
\usepackage{url}
\usepackage[ruled,vlined,linesnumbered,noresetcount]{algorithm2e}
 \usepackage{algpseudocode}

 \newlength\fwidth

 \usepackage{pgfplots}
\pgfplotsset{compat=newest}
\usepackage{tikz}
\usetikzlibrary{positioning}
\usetikzlibrary{shapes.geometric}
\usetikzlibrary{shapes.misc}
\usetikzlibrary{external}
\usetikzlibrary{shapes,arrows,patterns}
\usetikzlibrary{arrows.meta}
\usetikzlibrary{plotmarks}
\usetikzlibrary{calc,decorations.markings,math}

\usepackage{placeins} 
\usepackage{tabularx}

\graphicspath{ {pictures/} }
\newacronym{ap}{AP}{Average Precision}
\newacronym{av}{AV}{Autonomous Vehicles}
\newacronym{av2}{AV2}{Argoverse2}
\newacronym{aws}{AWS}{Amazon Web Services}
\newacronym{bev}{BEV}{bird's eye view}
\newacronym{cbgs}{CBGS}{Class-balanced grouping and sampling}
\newacronym{ccl}{CCL}{Connected Component Labelling}
\newacronym{cds}{CDS}{Composite Detection Score}
\newacronym{dnn}{DNN}{Deep Neural Network}
\newacronym{ekf}{EKF}{Extended Kalman Filter}
\newacronym{fsd}{FSD}{Fully Sparse Detector}
\newacronym{fsf}{FSF}{Fully Sparse Fusion}
\newacronym{gps}{GPS}{Global Positioning System}
\newacronym{hd}{HD}{High Definition}
\newacronym{htc}{HTC}{Hybrid Task Cascade}
\newacronym{kf}{KF}{Kalman Filter}
\newacronym{lidar}{LiDAR}{Light Detection and Ranging}
\newacronym{map}{mAP}{mean Average Precision}
\newacronym{mav}{MAV}{Micro Aerial Vehicle}
\newacronym{mhe}{MHE}{Moving Horizon Estimation}
\newacronym{mlp}{MLP}{Multi Layer Perceptrons}
\newacronym{mvp}{MVP}{Multimodal Virtual Points}
\newacronym{nds}{NDS}{NuScenes Detection Score}
\newacronym{nmhe}{NMHE}{Nonlinear Moving Horizon Estimation}
\newacronym{nms}{NMS}{Non-Maximum Suppression}
\newacronym{pdf}{PDF}{Probability Density Function}
\newacronym{radar}{RADAR}{Radio Detection and Ranging}
\newacronym{rbgs}{RBGS}{Range-balanced grouping and sampling}
\newacronym{rpn}{RPN}{Region Proposal Networks}
\newacronym{sir}{SIR}{Sparse Instance Recognition}
\newacronym{slam}{SLAM}{Simulataneous Localization and Mapping}
\newacronym{soa}{SoA}{State of the Art}
\newacronym{sota}{SOTA}{state-of-the-art}
\newacronym{sph}{SPH}{sparse prediction head}
\newacronym{vvm}{VVM}{Virtual Voxel Mixer}

\usepackage{amsmath} 
\usepackage[nameinlink,capitalise]{cleveref}
\usepackage{subcaption} 

\title{\LARGE \bf
Addressing Data Annotation Challenges in Multiple Sensors: A Solution for Scania Collected Datasets
}

\author{Ajinkya Khoche$^{1,2}$, Aron Asefaw$^{1}$, Alejandro González$^{2}$, Bogdan Timus$^{2}$, Sina Sharif Mansouri$^{2}$ \\ and Patric Jensfelt$^{1}$
\thanks{*This work was supported by the research grant PROSENSE (2020-02963) funded by VINNOVA. 
}
\thanks{$^{1}$KTH Royal Institute of Technology, Stockholm 10044, Sweden. Corresponding author's e-mail: {\tt\small khoche@kth.se}}%
\thanks{$^{2}$Autonomous Transport Solutions Lab, Scania Group, Södertälje, SE-15139, Sweden}
}

\begin{document}


\maketitle
\thispagestyle{empty}
\pagestyle{empty}

\begin{abstract}

Data annotation in autonomous vehicles is a critical step in the development of \gls{dnn} based models or the performance evaluation of the perception system. This often takes the form of adding 3D bounding boxes on time-sequential and registered series of point-sets captured from active sensors like \gls{lidar} and \gls{radar}. When annotating multiple active sensors, there is a need to motion compensate and translate the points to a consistent coordinate frame and timestamp respectively. However, highly dynamic objects pose a unique challenge, as they can appear at different timestamps in each sensor's data. Without knowing the speed of the objects, their position appears to be different in different sensor outputs. Thus, even after motion compensation, highly dynamic objects are not matched from multiple sensors in the same frame, and human annotators struggle to add unique bounding boxes that capture all objects. This article focuses on addressing this challenge, primarily within the context of Scania-collected datasets. The proposed solution takes a track of an annotated object as input and uses the \gls{mhe} to robustly estimate its speed. The estimated speed profile is utilized to correct the position of the annotated box and add boxes to object clusters missed by the original annotation. 
\end{abstract}

\glsresetall

\section{Introduction} 
\label{sec:intro}

\glsunset{lidar}
\glsunset{radar}

The pursuit of autonomous vehicles, particularly in heavy vehicle manufacturing, has gained momentum across various industries. At its core, the essential element driving the progress is ground truth data, essential for evaluating the \gls{av} software stack. To obtain this invaluable data, several approaches have emerged. One method involves equipping the ego and multiple non-ego vehicles with \gls{gps} sensors and orchestrating staged scenarios, for instance, overtaking, U-turns, roundabouts etc. While the \gls{gps} enables comprehensive state-awareness of the environment and provides valuable insights, this approach can't be extended to real-world driving. As such, highly controlled scenarios also fall short in emulating the complexities of real-world driving, leaving gaps in the generalization ability of the system. 

\glspl{dnn} have recently emerged as the backbone of autonomous driving systems, with various approaches proposed throughout the \gls{av} stack, including for perception~\cite{khoche2022semantic,khoche2023fully}, localization~\cite{chen2023deep}, motion prediction and situational awareness~\cite{seff2023motionlm}, control and path planning~\cite{teng2023motion}, vehichle-to-vehicle communication~\cite{li2022v2x} as well as end-to-end driving systems~\cite{hu2023planning}. A majority of \gls{dnn} approaches are trained using supervised learning and require annotations.
Simulation systems offer another avenue for generating training data at scale. However, the challenge lies in bridging the gap between simulated and real-world sensory output~\cite{hu2023simulation}. The successful integration of synthetic data into practical experiments remains a persistent question in the journey toward autonomous vehicles. The third approach, albeit time-consuming and expensive, involves annotating datasets collected from vehicles on the road. This method necessitates rigorous quality checks.

\begin{figure}[tbp!] 
\centering
\includegraphics[width=0.85\linewidth]{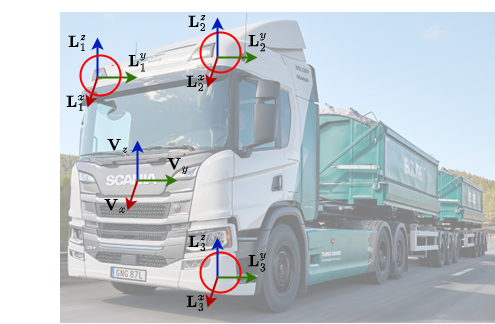}
\caption{Illustration of the Scania truck with different sensor placement highlighted with red circles, while the $i^{th}$ sensor and the vehicle coordinate frame are shown as $\mathbf{L_i}$ and $\mathbf{V}$ respectively.}
\label{fig:concept}
\end{figure}

The field of autonomous driving has seen an abundance of datasets capturing challenging real driving scenarios. The KITTI dataset, proposed by Geiger et. al.~\cite{geiger2012we}, was a pioneering work in this regards, providing annotations for 3D object detection, stereo matching and optical flow, as well as high quality position labels to enable research in \gls{slam}. 
The NuScenes~\cite{caesar2020nuscenes}, Waymo~\cite{sun2020scalability} and Argoverse~\cite{chang2019argoverse} datasets are notable for their large scale and diversity, including night-time driving and adverse weather scenarios.
Many datasets also provide access to \gls{hd} maps to enable advanced processing~\cite{chang2019argoverse,caesar2020nuscenes}.
The recently proposed Argoverse2~\cite{wilson2023argoverse}, aiMotive~\cite{matuszka2022aimotive} and Zenseact~\cite{alibeigi2023zenseact} datasets provide annotations focusing on long range perception. 


\begin{figure}[htbp!] 
\centering
\includegraphics[width=0.75\linewidth]{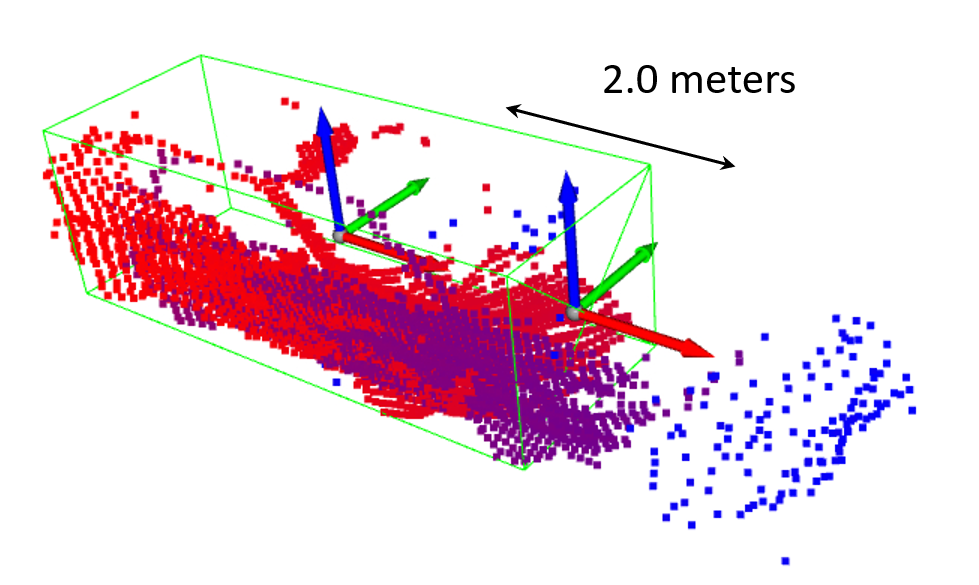}
\caption{Point cloud from three sensors after compensating for ego motion, (red, violet and blue points are from \gls{lidar}s $L_1$, $L_2$ and $L_3$ respectively). The dynamic object in the scene is observed by different sensors in different time stances, with displacement of 2~\unit{m} within 100~\unit{ms}. The manually annotated 3D bounding box is shown in green.}
\label{fig:annotation}
\end{figure}

However most of the existing datasets are captured onboard passenger cars, and their characteristics differ significantly from the data captured onboard trucks. 
Due to their larger size compared to passenger cars, trucks require a greater number of sensors with increased spacing between them to provide comprehensive surround views, as shown in \Cref{fig:concept}. This extended displacement between the sensors leads to them capturing different, often non-overlapping views of the same object. 
Moreover, the available datasets mostly consist of suburban scenarios with low driving speeds, whereas long haulage trucks operate on highways, with objects moving at high speeds. These dynamic objects are captured at multiple positions by various sensors, even after ego-motion compensation. To the best of our knowledge, none of the existing datasets capture this phenomenon.

The aforementioned issues pose notable challenges for human annotators when attempting to define accurate 3D bounding boxes around objects. Firstly, the human-labelled bounding boxes may not encompass the entirety of the point clouds, leading to scenarios where portions of the objects, as illustrated in \Cref{fig:annotation}, remain outside the bounding box's scope. Secondly, the annotators might label different views of an object at various time instances, leading to inaccurate 
speed estimation of the vehicle. These annotations, if not refined may lead to an incorrect evaluation of perception algorithms, or be a source of error during training of \gls{dnn} models. 

In this work, we address this problem by modelling the annotated boxes as noisy measurements of the object state. Consequently, state estimation algorithms can be used to infer the object's true state. Given an annotated object track from a multi-\gls{lidar} dataset as noisy positional inputs, this article proposes using \gls{mhe} as a state estimator to predict the position and speed of non-ego objects. The estimates are subsequently used to  refine the positioning of bounding box annotations to cover all the views of the object.

\subsection{Background \& Motivation}

\gls{kf}~\cite{kalman1960new} is widely used in applications where the system dynamics and measurement models are both linear and the noise is Gaussian, while \gls{ekf} can handle non-linear system and measurement models. \gls{kf} based estimators may have slow convergence for rapid changes in state, and only consider one measurement for each estimation iteration. \gls{nmhe} methods are also getting more attention~\cite{haseltine2005critical,papadimitriou2022external,mansouri2020external} for their ability to estimate complex nonlinear dynamic models, while they can handle inequality constraints. \gls{mhe} method uses a moving time window to iteratively estimate the states of a nonlinear dynamic system, providing real-time updates as new measurements become available. The \gls{mhe}, driven by its optimization-based framework and its ability to utilize a set of measurements, is a preferred choice for accurately estimating the speed of non-ego vehicles in various scenarios, with the added advantage of being able to handle constraints such as bounds on vehicle speed, making it a versatile solution.


\subsection{Contributions}
With the abovementioned state-of-the-art as the context, the key contributions of the article are provided in this section. The first and foremost contribution is to highlight the problem of annotation for multiple active sensors in heavy vehicles, which to the best of authors' knowledge, has not been discussed before. As the second contribution, a \gls{mhe} estimator with kinematic model is formulated to estimate the non-ego vehicle speeds, which is used to further rectify the bounding box annotations.
The third contribution stems from the experimental evaluation of the method on a \gls{lidar} dataset captured onboard trucks and buses containing diverse scenarios of non-ego object motion.

\subsection{Outline}
The rest of the article is organized as follows: \Cref{sec:prob_statement} outlines the problem. The \gls{mhe} formulation is described in \Cref{sec:method}. \Cref{sec:expt}
sets up the experiments and discusses results on data captured onboard Scania platform. Lastly 
\Cref{sec:conclusions} concludes the article by summarizing the findings and discussing directions for future work. 

\section{Problem Statement} \label{sec:prob_statement}


\begin{figure*}[tbp!]
\centering
\includegraphics[width=0.9\linewidth]{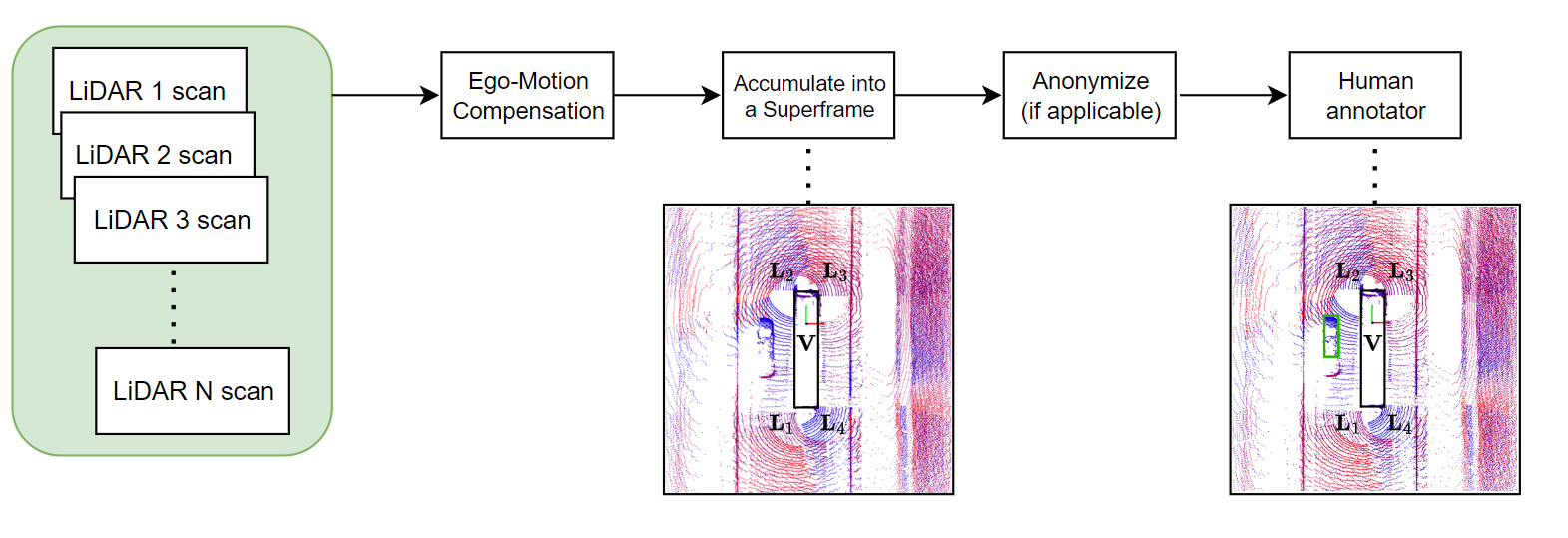}
\caption{The annotation process for Scania collected dataset. Scans from multiple \gls{lidar}s are motion compensated and accumulated in a \textit{superframe}. Thereafter a time-series of superframes are post processed and sent for manual annotation. A snapshot of a superframe before and after annotation is shown. The ego vehicle is marked in the center with vehicle frame $\mathbf{V}$, and sensor frames $\mathbf{L}_1$ to $\mathbf{L}_4$. The point clouds from multiple \gls{lidar}s are colored according to their offset from motion compensated timestamp (chosen to be the middle of the superframe). Red and blue indicate beginning and end of the superframe respectively. The annotated vehicle is shown with a green box.}
\label{fig:prob_statement}
\end{figure*} 

Let $\mathbf{W}$, $\mathbf{V_t}$ and $\mathbf{L^i_t}$ represent the world frame, the vehicle frame at time $t$, and the $i^{th}$ sensor frame at time $t$ respectively. $^{W}\mathbf{T}_{V_t}$ denotes the $4\times4$ homogeneous transformation matrix representing the vehicle's pose in the world frame at time $t$. Let $^{L^i_t}\mathbf{P}_i(t)$
denote a point cloud acquired by the $i^{th}$ \gls{lidar} sensor during the time period $[t, t+\Delta{t}]$. The $j^{th}$ point in this point cloud is represented by a 3D position $p_{i,j}=(x_{i,j}, y_{i,j}, z_{i,j})^T$ in the sensor frame and timestamp $\tau_{i,j}$.
Since the ego-vehicle is moving continuously while the rotating \gls{lidar} captures data, a transformation needs to be applied to each point of the point cloud, to compensate for the ego-motion to simplify processing by deskewing the point cloud. 
Let $t^*$ denote the time to which the motion is compensated. In our work $t^* = t + \Delta{t}/2$, i.e. in the center of the scan period of the point cloud.
We denote by ${}^{V_{t^*}}\mathbf{P}_i^{*}$ the motion compensated point cloud where the position, $p_{i,j}$, of each point is transformed to the vehicle frame at time $t^*$. The motion compensated points are denoted ${}^{V_{t^*}}p_{i,j}^*$, which for brevity will be written as $p_{i,j}^*$ and are calculated as:
\begin{subequations} \label{eq:motion_compensation}
    \begin{align} 
        p_{i,j}^* = {}^{V_{t^*}}\mathbf{T}_{V_{\tau_{i,j}}} {}^{V_{\tau_{i,j}}}\mathbf{T}_{L^i_{\tau_{i,j}}} p_{i,j}  \\
        {}^{V_{t^*}}\mathbf{T}_{V_{\tau_{i,j}}} = ({}^{W}\mathbf{T}_{V_{{t}^*}})^{-1} {}^{W}\mathbf{T}_{V_{\tau_{i,j}}}
    \end{align}
\end{subequations} 
where ${}^{V_{\tau_{i,j}}}\mathbf{T}_{L^i_{\tau_{i,j}}}$ is given by the extrinsic calibration and is here assumed constant ${}^{V}\mathbf{T}_{L^i}$. A constant linear and angular velocity is assumed in the interval $[t,t+\Delta{t}]$. 

The focus of our work is a multi-\gls{lidar} setup. We define a \textit{superframe} ${}^{V_{{t}^*}}\mathbf{P}_{S}$ as a point cloud accumulating all motion compensated points from $M$ \gls{lidar}s within a time interval $[t,t+\Delta{t}]$,
\begin{align}
    {}^{V_{{t}^*}}\mathbf{P}_{S} = \bigcup_{i=1}^{M} {}^{V_{{t}^*}}\mathbf{P}_i^*\label{eq:superframe}.
\end{align} 

A time-series of motion compensated superframes is post-processed and sent for annotation of 3D boxes, as shown in \Cref{fig:prob_statement}. Notably, \cref{eq:motion_compensation} follows a static world assumption, which does not hold in real-world driving applications. For example, for $\Delta{t}=100~\unit{ms}$, a non-ego vehicle driving at 30~\unit{m/sec} on a highway could have a worst case displacement of up to 3~\unit{meters} captured by different sensors within the superframe. A single bounding box is inadequate in capturing this motion, as shown in \cref{fig:prob_statement}. 
The problem then involves modeling the motion of the non-ego object given noisy measurements of the state provided by a time series (or a track) of annotated 3D bounding boxes.


An agent's 3D motion can be generally described by 12 states: $\bm{x}=[x,y,z,\Dot{x},\Dot{y},\Dot{z},\phi,\gamma,\theta,\Dot{\phi},\Dot{\gamma},\Dot{\theta}]^{T}$ where the first six states denote the positions and linear velocities, and the last six states denote the roll, pitch and yaw angles and their rates respectively.
Particularly for driving scenarios, a planar motion with holonomic constraints is considered. This assumption removes the need to estimate the $z$ position, the roll and pitch angles, as well as their rates, reducing the state space to six.
Furthermore, assuming the measured heading of an object 
remains constant within the superframe interval $\Delta{t}$, and the heading is error-free, the problem can be further simplified to a one-dimensional estimation containing two states $\bm{x}=[d,s]^{T}$, where $d$ and $s$ are the distance and speed along a specified trajectory.

\section{Methodology} \label{sec:method}
In this section, we will present our \gls{mhe} approach. We will begin by outlining the mathematical formulation that underlies our method, offering an explanation of the equations and principles. Subsequently, we will explain the estimation of non-ego vehicle speeds in \gls{mhe}. Finally, we will present the architecture that rectifies annotations, providing a holistic view of our approach's practical implementation.

\subsection{Kinematic Model} \label{subsec:model}
Given the state $\bm{x}=[d, s]^{T}$ as described in the previous section, the distance $d$ is specified
along a trajectory specified by a set of error-free headings $\bm{\Theta} = \lbrace \theta_i : i = 1, ... ,n_l \rbrace$, where $n_l$ is the length of an annotated track. The measurements here come from human annotations.
The speed $s$
is not measured. The state of a non-ego vehicle at time $t$ is assumed to be given by the constant acceleration model \eqref{eq:modelvehicle}.
\begin{subequations}
    \label{eq:modelvehicle}
    \begin{align}
    d(t) &= d_0 + s(t) t + \frac{1}{2} a t^2, \\
    s(t) &= s_0 + a t.
    \end{align}
    \label{eq:kinematic_model}
\end{subequations}
In this work we study short trajectories. A more general solution would need to adopt a more complex motion model.

\subsection{Moving Horizon Estimation} \label{subsec:MHE}

The main objective of the \gls{mhe}~\cite{rao2003constrained} is to obtain the state estimate at time $t$, given a set of measurements collected over a moving horizon of past time steps. The state and measurement are modelled in discrete form as:
\begin{subequations}\label{eq:discretizationmodel}
\begin{align}
\bar{\bm{x}}_{k+1} &= \mathcal{F}(\bar{\bm{x}}_k,\bm{u}_{k})+ \bm{w}_k,\\
\bm{y}_k &= \mathcal{H}(\bar{\bm{x}}_{k}) + \bm{\Lambda}_{k},
\end{align}
\end{subequations}
where, $\bar{\bm{x}}_{k+1}$ is the state estimate at time step $k+1$, $\mathcal{F}: \mathbb{R}^{n_s} \times \mathbb{R}^{n_u} \to \mathbb{R}^{n_s} $ is the state transition function and $\bm{u}_{k}$ is the control input at time step k. $\bm{u}_{k}$ is the same as the acceleration $a$. $\bm{y_k}$ is the modelled measurement and $\mathcal{H}: \mathbb{R}^{n_s} \to \mathbb{R}^{n_m} $ is the measurement function. Moreover $n_s$, $n_u$, and $n_m$ are the number of states, inputs and measurements respectively, $\bm{\Lambda}_{k} \in \mathbb{R}^{n_m}$ and $\bm{w}_{k} \in \mathbb{R}^{n_s}$ represent the measurement noise and the process noise correspondingly. The initial state estimate $\bar{\bm{x}}_0$ is known. Furthermore, $\bar{\bm{x}}_{k-j|k}$ and $\bm{y}_{k-j\mid k}$ are the previous $k-j$ state and measurements as observed from the current time $k$.

The process noise $\bm{w}_{k}$, measurement noise $\bm{\Lambda}_{k}$ and the initial state estimate noise are unknown and assumed to follow a random distribution, characterized by the Gaussian \gls{pdf} with the covariance $\bm{Q} \in \mathbb{R}^{n_s \times n_s}$, $\bm{\Omega} \in \mathbb{R}^{n_m \times n_m}$, and $\bm{\Psi} \in \mathbb{R}^{n_s \times n_s} $ respectively~\cite{ungarala2009computing}. 
%

Given a set of noisy measurements $\bm{Y} = \lbrace \bm{y}_j : j = k-N_e, ... ,k-1 \rbrace$ in a fixed horizon window $N_e$ , the estimated states of the system $\bar{\bm{X}}\ = \lbrace \bar{\bm{x}}_j : j = k-N_e, ... ,k-1 \rbrace $ are obtained by solving the optimization problem:

\begin{subequations}\label{eq:mhe1}
\begin{align}
&\min_{\bar{\bm{x}}_{(k-N_e \mid k)},\bm{W}_{(k-N_e \mid k)}^{(k-1 \mid k)}}    {J(k)}\\
&\textrm{s.t.} \,\,  \bar{\bm{x}}_{i+1|k} = \mathcal{F}(\bar{\bm{x}}_{i|k},\bm{u}_{i|k})+ \bm{w}_{i|k}\\
&\bm{y}_{i|k} = \mathcal{H}(\bar{\bm{x}}_{i|k}) + \bm{\Lambda_{i|k}}  \quad i=\{ k-N_e, \dots k-1 \}  \\
& \bm{w}_k \in \mathbb{W}_k ,\quad \bm{\Lambda}_k\in \mathbb{\Lambda}_k ,\quad \bar{\bm{x}}_k \in \mathbb{X}_k
\end{align}
\end{subequations}
where,
\begin{equation}
\begin{aligned}\label{eq:mhe2}
&J(k) = 
    \underbrace{  
    (\bar{\bm{x}}_{k-N_e|k} - {\tilde{\bm{x}}}_{k-N_e|k})^{T}   \bm{\Psi}^{-1} (\bar{\bm{x}}_{k-N_e|k} - {\tilde{\bm{x}}}_{k-N_e|k})
    }_\text{arrival cost} \\ &
  + \sum_{i=k-N_e}^{i = k}  
    \underbrace{ 
    (\bm{y}_{i|k} - \mathcal{H}(\bar{\bm{x}}_{i|k}))^{T} \bm{\Omega}^{-1} (\bm{y}_{i|k} - \mathcal{H}(\bar{\bm{x}}_{i|k})) 
    }_{\text{stage cost}}   
    + \sum_{i=k-N_e}^{i = k-1} \\ &
        \underbrace{ 
        (\bar{\bm{x}}_{i+1|k} - \mathcal{F}(\bar{\bm{x}}_{i|k},\bm{u}_{i|k}))^{T}  
        \bm{Q}^{-1} (\bar{\bm{x}}_{i+1|k} - \mathcal{F}(\bar{\bm{x}}_{i|k},\bm{u}_{i|k})) }_{\text{stage cost}} 
\end{aligned}
\end{equation}


In~\eqref{eq:mhe1} $\bm{W}_{(k-N_e)}^{(k-1)} = col (\bm{w}_{(k-N_e)}, \dots,\bm{w}_{(k-1)} )$ is the estimated process disturbance from time $k-N_e$ up to $k-1$, estimated at the time $k$. The first term of the objective $J$ is referred to as the arrival cost. It measures the squared difference between the current and the prior state estimate at the beginning of the horizon window. In effect, the arrival cost is a mechanism for incorporating historical state information into the current estimation problem, ensuring a smooth transition from past estimates to current estimates. 
The remaining terms are denoted as stage cost or incremental cost. They compute the sum of the squared difference between the predicted and modelled measurement, and the predicted and modelled state respectively. The predictions come from the measurement function and the state transition function. Additionally, the terms are weighted by covariances of the inverse of initial state estimate, measurement and the process noise respectively. A smaller covariance indicates higher confidence in the previous estimate, leading to a larger penalty for deviations. A finite-horizon optimal problem with horizon window of $N_e$ is solved at every time step $k$, to obtain the corresponding state estimates $\bar{\bm{x}}_{k-N_e|k}^{\star},\dots$ $\bar{\bm{x}}_{N_e|k}^{\star}$. 


%
\subsection{Refining Multi-\gls{lidar} Annotations} \label{subsec:annotation_refinement}
\begin{figure*}[tbp!] 
\centering
\includegraphics[width=0.9\linewidth]{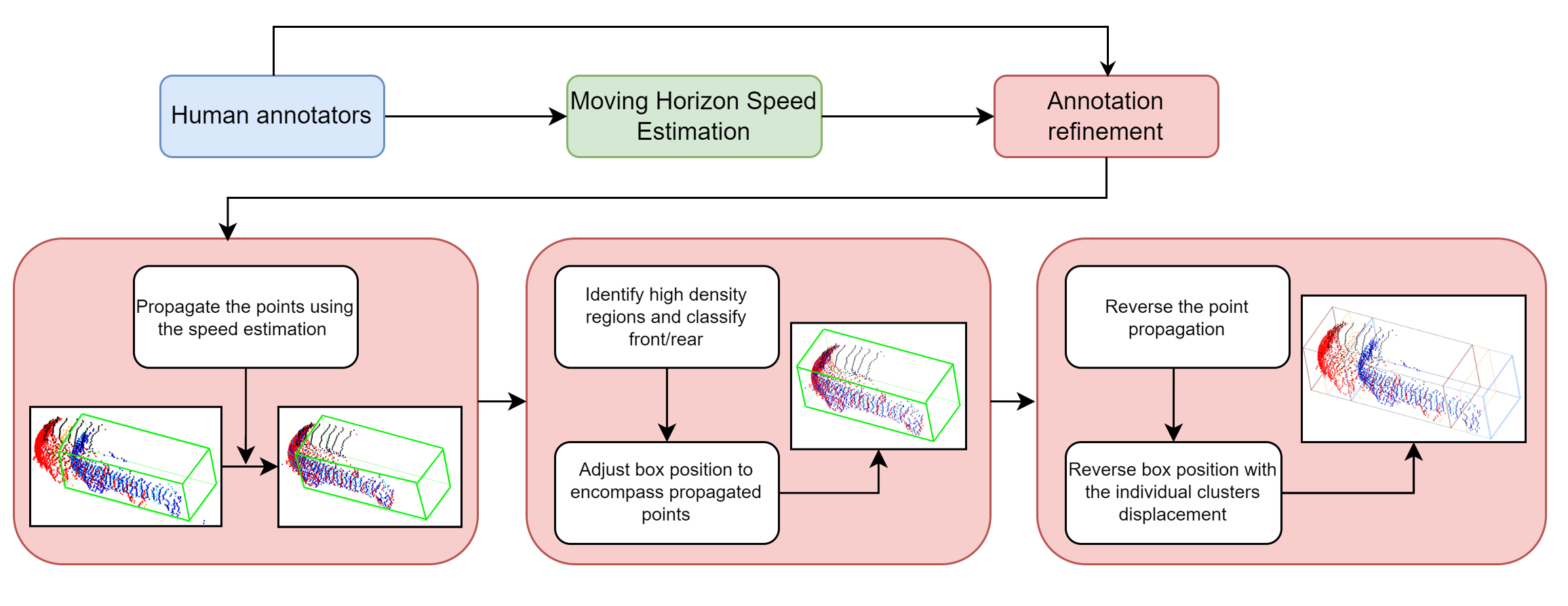}
\caption{Proposed approach for refining multi-\gls{lidar} annotations. 
The input is time-sequential track of human labelled annotation (shown with green box) alongside it's speed estimate obtained using \gls{mhe}. 
Clustering along heading captures red points, which represent a view of the vehicle missed by the human annotated box. 
Next, the \gls{mhe} speed estimate is used to shift the points according to the difference between $\tau_{i,j}$ and $t^*$. The red points move forward, whereas the blue points move back. Thirdly, the annotated box is moved to align with the shifted points. Lastly, the green box is duplicated for each cluster, and both the points and pseudo bounding boxes are shifted back according to the reverse cluster displacement. The pseudo bounding boxes are colored according to their best fitting clusters.   
}
\label{fig:blockfiagramofsolution}
\end{figure*}

The goal of our work is to refine the manual annotations for each non-ego vehicle. We do this by generating boxes corresponding to each individual \gls{lidar}, given a single annotation per non-ego object for each superframe.
Solving for \Cref{eq:mhe1} traversing along the horizon window $N_e$ provides the optimal state estimate for the entire track, denoted as $\bar{\bm{X}}^{\star}_{1:n_m}$.
These estimates, in conjunction with the human-annotated bounding boxes, are utilized as the input for box refinement, as shown in \Cref{fig:blockfiagramofsolution}.
The first step in this approach involves clustering along the box heading to get $G$ different views of the object captured by various sensors within a superframe. As the second step, the estimated speed $s_k^{\star}$ is used alongside the box heading $\theta_k$ to 
compensate all points classified as being part of the non-ego vehicle for its speed. Concretely, for all \gls{lidar}'s $i$ and points $j$ that correspond to the non-ego vehicle we calculate the change in position as:


\begin{subequations} \label{eq:delta_p}
\begin{align} 
    \Delta p_{i,j} = 
    \begin{bmatrix}
        \Delta x_{i,j} \\ \Delta y_{i,j} \\ \Delta z_{i,j}
    \end{bmatrix} =
    \begin{bmatrix}
        (\tau_{i,j} - t^{\star}) s_k^{\star} \cos \theta_k \\
        (\tau_{i,j} - t^{\star}) s_k^{\star} \sin \theta_k \\
        0
    \end{bmatrix} \\
\end{align}
\end{subequations}

\noindent Where $\tau_{i,j}$ is the timestamp of point $p_{i,j}$, and $t^*$ is the motion compensation time. $\Delta p_{i,j}$ is added to $p_{i,j}$ to obtain speed-compensated points for the non-ego vehicle.
Thereafter the front or rear of the vehicle are inferred using the highest density region~\cite{parzen1962estimation} along the heading. 
If the high density region lies behind the euclidean mean, the region represents the back of the object, else the front. This knowledge allows the algorithm to position the bounding box by anchoring the edges to the extreme points of the vehicle along the direction of travel.
Next, $G$ copies of the bounding box are produced, one for each cluster. These are denoted as \textit{pseudo bounding boxes}. Lastly, these copies are each shifted back by the inverse of $\Delta p_{i,j}$, leading to the pseudo boxes fitting precisely with the original clusters. 

\section{Experiments}   \label{sec:expt}
In this section, we present our experimental setup, as well as discuss results of the state estimation using \gls{mhe} and the annotation refinement. In this article, we focus on highly dynamic vehicles as the problem is more pronounced for such cases. But our approach could be extended to cover other classes eg. articulated vehicles, pedestrians etc. as well.

\subsection{Experimental Setup}   \label{subsec:expt_setup}
Sequences of multi-\gls{lidar} data was collected and annotated on Scania platforms consisting of trucks and buses. The dataset encompasses a wide range of scenarios, including urban and highway driving, as well as challenging adverse weather conditions.
The annotated sequences have a duration of 10~\unit{s} and include motion-compensated \gls{lidar} points captured at 10~\unit{FPS}, 3D bounding boxes, class labels, and tracking IDs for each object. The annotators utilized keyframes to extract essential parameters, such as class and size. Keyframes are selected based on the time sequence, specifically focusing on the frames that capture the highest quality representation of the objects in that sequence. It's important to note that this selection may vary for different objects. It's worth mentioning that, for the sake of generality, we did not account for this variability in our work.

For state estimation using \gls{mhe}, the kinematic model described in \Cref{subsec:model} is utilized as the state transition model $\mathcal{F}$. 
Since the problem at hand is solved offline, we chose the horizon window $N_e$ to be the same as entire length of measurements, i.e. length of an annotated track.
The state estimates of the entire track are thus optimized in a single iteration. 
The \gls{mhe} parameters are indicated as $n_s=2$ and $n_m=1$, as only the distance is measured.  $\bm{\Psi}^{-1}=\mathbf{I}_{2 \times 2}$, $\bm{Q}=\mathbf{I}_{2 \times 2}$, and $\bm{\Omega}=\mathbf{I}_{1 \times 1}$ where $\mathbf{I}$ is identity matrix. 

\subsection{Results}  \label{subsec:results}
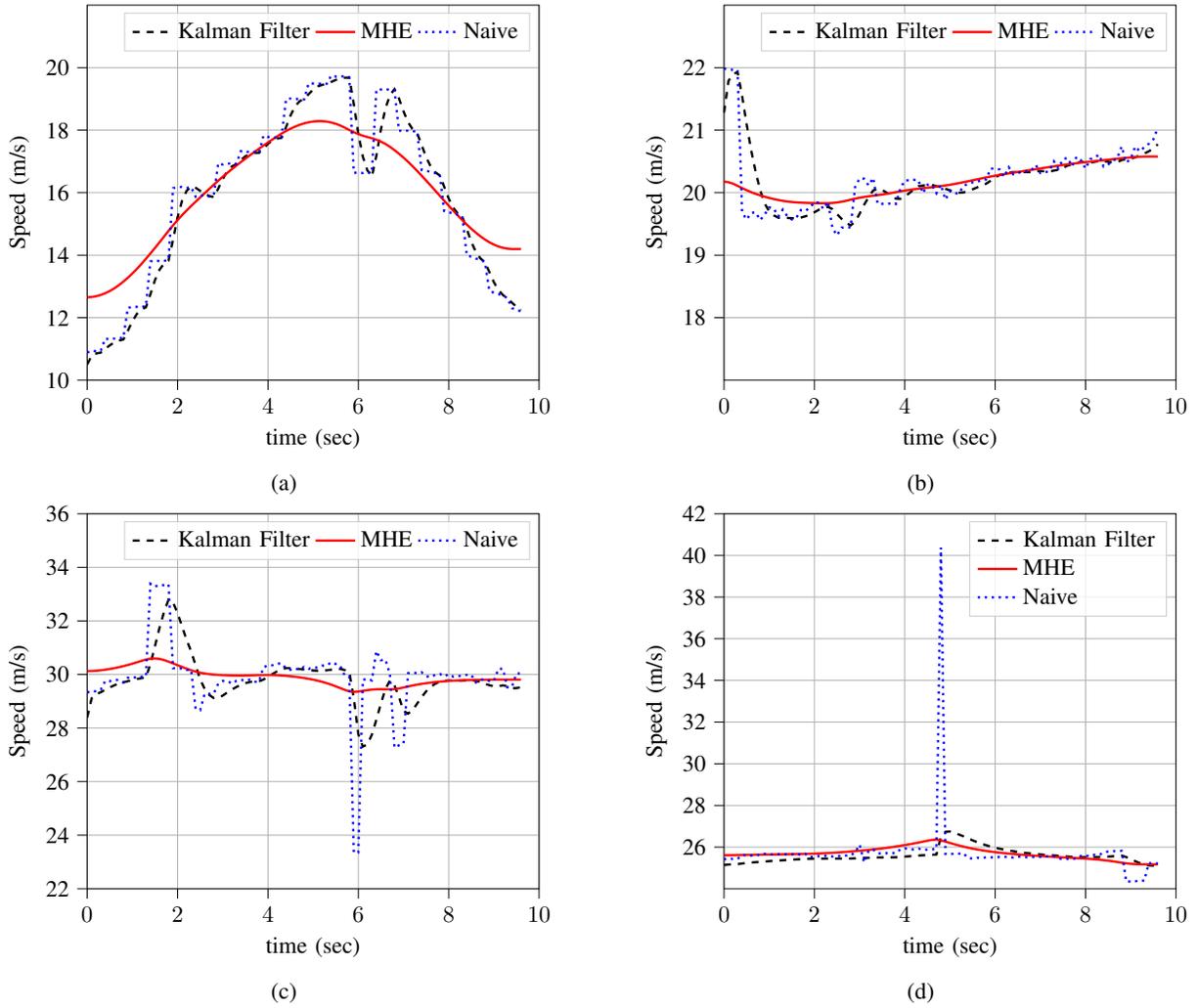
\begin{figure*} 
    \centering
    \begin{subfigure}[b]{0.48\textwidth}
        \centering
        \resizebox{0.9\textwidth}{!}{
\begin{tikzpicture}

\definecolor{darkgray176}{RGB}{176,176,176}
\definecolor{darkorange25512714}{RGB}{255,127,14}
\definecolor{forestgreen4416044}{RGB}{44,160,44}
\definecolor{lightgray204}{RGB}{204,204,204}
\definecolor{steelblue31119180}{RGB}{31,119,180}

\begin{axis}[
legend cell align={left},
legend style={fill opacity=0.8, draw opacity=1, text opacity=1, draw=lightgray204, legend columns=3},
tick align=outside,
tick pos=left,
x grid style={darkgray176},
xlabel={time (sec)},
xmajorgrids,
xmin=0, xmax=10,
xtick style={color=black},
y grid style={darkgray176},
ylabel={Speed (m/s)},
ymajorgrids,
ymin=10, ymax=22,
ytick style={color=black},
ytick={10,12,14,16,18,20},
yticklabels={10,12,14,16,18,20}
]
\addplot [color=black,dashed, line width= 1.0 pt]
table {
0 10.5020581152715
0.1 10.7903761376943
0.2 10.8538236340768
0.3 10.8818527865196
0.4 11.0008444493314
0.5 11.1154243666671
0.6 11.2067375514281
0.7 11.266612620647
0.8 11.3002682579074
0.9 11.5799418530793
1 11.885374302389
1.1 12.1197325069766
1.2 12.264240631187
1.3 12.3334285093335
1.4 12.7512598719414
1.5 13.2000087164248
1.6 13.5321016898424
1.7 13.7292686611346
1.8 13.8268974346664
1.9 14.4899267042819
2 15.205200919043
2.1 15.7442986866271
2.2 16.0568629705705
2.3 16.2021589440315
2.4 16.155814524827
2.5 16.0576388436642
2.6 15.9659694528373
2.7 15.8933965044165
2.8 15.8570024153272
2.9 16.1294689384784
3 16.437561791939
3.1 16.6837676765221
3.2 16.8377238779951
3.3 16.9100543895735
3.4 17.0425855972692
3.5 17.1624646100165
3.6 17.2324582102309
3.7 17.2697871162055
3.8 17.2719414411253
3.9 17.4144580459711
4 17.5612767004757
4.1 17.6731317989348
4.2 17.7282975233723
4.3 17.7551439101673
4.4 18.0955620054443
4.5 18.4729823648553
4.6 18.7584203284109
4.7 18.9261561235126
4.8 18.998399586751
4.9 19.1526220842612
5 19.2968404992555
5.1 19.3974127833959
5.2 19.4502809567647
5.3 19.4697226783349
5.4 19.542377949279
5.5 19.6076183281191
5.6 19.6519726517567
5.7 19.6782561212625
5.8 19.6821147327202
5.9 18.8351371971505
6 17.8845772079284
6.1 17.1518826618015
6.2 16.7143233296041
6.3 16.5093279196786
6.4 17.1827746906002
6.5 18.0212370455375
6.6 18.6982662824946
6.7 19.1154510972077
6.8 19.3122495947318
6.9 19.0134179043782
7 18.6007050999325
7.1 18.2565087254174
7.2 18.0286490361707
7.3 17.9095331488878
7.4 17.525819706278
7.5 17.134749398733
7.6 16.840134991889
7.7 16.6537634142365
7.8 16.5532586842221
7.9 16.188591027698
8 15.8059372986098
8.1 15.5188322024271
8.2 15.3319819255001
8.3 15.2252200219306
8.4 14.8328855747239
8.5 14.4245121161691
8.6 14.1056702439871
8.7 13.8989933235411
8.8 13.7750282071748
8.9 13.457123078211
9 13.1334558979454
9.1 12.8855736493687
9.2 12.7221414693392
9.3 12.619652352867
9.4 12.4796820829777
9.5 12.3468889771211
9.6 12.2424733951973
};
\addlegendentry{Kalman Filter}
\addplot [color=red,  line width= 1.0 pt]
table {%
0 12.6507820960786
0.1 12.6613025608338
0.2 12.6863081624062
0.3 12.7268833185599
0.4 12.7822639497296
0.5 12.8525456578148
0.6 12.9378774641921
0.7 13.0383248277184
0.8 13.152980816094
0.9 13.2785537189247
1 13.414267886241
1.1 13.5604111738738
1.2 13.7173101448841
1.3 13.883998473888
1.4 14.0555499815617
1.5 14.2308690770581
1.6 14.410165951287
1.7 14.5932131141826
1.8 14.7775057643944
1.9 14.9543964671142
2 15.1215887496748
2.1 15.2793200921773
2.2 15.4288079649247
2.3 15.5717000236849
2.4 15.7106074174208
2.5 15.8471229646218
2.6 15.982534994791
2.7 16.1177360728308
2.8 16.2522425065543
2.9 16.3827398158624
3 16.5088376269833
3.1 16.6309030080818
3.2 16.7496247561611
3.3 16.8654983177642
3.4 16.9778568279177
3.5 17.0871670011625
3.6 17.1943091213796
3.7 17.2998901415937
3.8 17.4041646969923
3.9 17.5057485745862
4 17.604802360743
4.1 17.7017693403822
4.2 17.7968192512046
4.3 17.8887000287644
4.4 17.9728266101808
4.5 18.0478076503969
4.6 18.1135214894055
4.7 18.1700336094022
4.8 18.2169696796573
4.9 18.2524700932884
5 18.2759663238217
5.1 18.2873482641499
5.2 18.2865004244685
5.3 18.2730194057615
5.4 18.2457262029412
5.5 18.2042963594822
5.6 18.148941326402
5.7 18.0810336018439
5.8 18.0052472358695
5.9 17.9345177503072
6 17.8731871840919
6.1 17.8220012323254
6.2 17.779366667764
6.3 17.7406241783026
6.4 17.6935831183141
6.5 17.6333181380907
6.6 17.5577597197477
6.7 17.4665009248356
6.8 17.3606996310173
6.9 17.2450940495073
7 17.1208684265062
7.1 16.9879780514535
7.2 16.8463401677293
7.3 16.6968483373613
7.4 16.5437269482293
7.5 16.3879018700336
7.6 16.2292233846334
7.7 16.0675342233996
7.8 15.9036945104943
7.9 15.7416929956936
8 15.5824123964643
8.1 15.4256090611134
8.2 15.2712644710282
8.3 15.1204768046259
8.4 14.9777075634121
8.5 14.8441763017559
8.6 14.7200951526869
8.7 14.6056153943816
8.8 14.5018130890918
8.9 14.4121676368871
9 14.3378377962746
9.1 14.2791281401185
9.2 14.2361245225066
9.3 14.2088878037869
9.4 14.1971355112046
9.5 14.1970628179075
9.6 14.1970197840389
};
\addlegendentry{MHE}
\addplot [color=blue,dotted, line width= 1.0 pt]
table {%
0 10.8847212791443
0.1 10.9448206424713
0.2 10.9255504608154
0.3 10.9454822540283
0.4 11.3320231437683
0.5 11.3011384010315
0.6 11.331775188446
0.7 11.3417565822601
0.8 11.3517332077026
0.9 12.317156791687
1 12.3371160030365
1.1 12.3470962047577
1.2 12.3570466041565
1.3 12.3562896251678
1.4 13.7993490695953
1.5 13.8301146030426
1.6 13.8079071044922
1.7 13.8174974918365
1.8 13.8481962680817
1.9 16.1399364471436
2 16.1819291114807
2.1 16.181173324585
2.2 16.1484456062317
2.3 16.1802363395691
2.4 15.8683466911316
2.5 15.9094703197479
2.6 15.9094703197479
2.7 15.8761262893677
2.8 15.9085786342621
2.9 16.9417238235474
3 16.8989777565002
3.1 16.9308817386627
3.2 16.931654214859
3.3 16.9209134578705
3.4 17.3289382457733
3.5 17.3290145397186
3.6 17.2849345207214
3.7 17.3074185848236
3.8 17.2647738456726
3.9 17.8247165679932
4 17.7813017368317
4.1 17.7939581871033
4.2 17.7505350112915
4.3 17.783180475235
4.4 18.9886152744293
4.5 19.0126836299896
4.6 19.0027678012848
4.7 18.9918720722198
4.8 18.9819180965424
4.9 19.4920563697815
5 19.4931519031525
5.1 19.4921314716339
5.2 19.4832527637482
5.3 19.482079744339
5.4 19.7481727600098
5.5 19.7145569324493
5.6 19.7156620025635
5.7 19.7271645069122
5.8 19.695440530777
5.9 16.6378664970398
6 16.6378676891327
6.1 16.6278910636902
6.2 16.6592621803284
6.3 16.6386294364929
6.4 19.2915034294128
6.5 19.3014812469482
6.6 19.3226647377014
6.7 19.2813766002655
6.8 19.27241563797
6.9 17.9814827442169
7 17.9825723171234
7.1 17.9823434352875
7.2 17.9626798629761
7.3 17.9615128040314
7.4 16.7274105548859
7.5 16.6974854469299
7.6 16.6677892208099
7.7 16.6268634796143
7.8 16.597090959549
7.9 15.4105567932129
8 15.3600001335144
8.1 15.3570985794067
8.2 15.2901077270508
8.3 15.2494585514069
8.4 13.9927065372467
8.5 13.9529085159302
8.6 13.8934051990509
8.7 13.8524436950684
8.8 13.7809872627258
8.9 12.8576397895813
9 12.7982914447784
9.1 12.75723695755
9.2 12.7078855037689
9.3 12.6240026950836
9.4 12.307094335556
9.5 12.2449731826782
9.6 12.2055172920227
};
\addlegendentry{Naive}
\end{axis}

\end{tikzpicture}}
        \caption{}
    \end{subfigure}
    \begin{subfigure}[b]{0.48\textwidth}
        \centering
        \resizebox{0.9\textwidth}{!}{
\begin{tikzpicture}

\definecolor{darkgray176}{RGB}{176,176,176}
\definecolor{darkorange25512714}{RGB}{255,127,14}
\definecolor{forestgreen4416044}{RGB}{44,160,44}
\definecolor{lightgray204}{RGB}{204,204,204}
\definecolor{steelblue31119180}{RGB}{31,119,180}

\begin{axis}[
legend cell align={left},
legend style={fill opacity=0.8, draw opacity=1, text opacity=1, draw=lightgray204, legend columns=3},
tick align=outside,
tick pos=left,
x grid style={darkgray176},
xlabel={time (sec)},
xmajorgrids,
xmin=0, xmax=10,
xtick style={color=black},
y grid style={darkgray176},
ylabel={Speed (m/s)},
ymajorgrids,
ymin=17, ymax=23,
ytick style={color=black},
ytick={18,19,20,21,22},
yticklabels={18,19,20,21,22}
]
\addplot [color=black, dashed,line width= 1.0 pt]
table {
0 21.2814243451156
0.1 21.7912918680789
0.2 21.8958984749239
0.3 21.9303701670629
0.4 21.5492166829844
0.5 21.0956433574647
0.6 20.6820145357863
0.7 20.3326009364844
0.8 20.0392701697946
0.9 19.8259612089939
1 19.7003866236608
1.1 19.6374171615459
1.2 19.5994177840453
1.3 19.5955492311324
1.4 19.595909412687
1.5 19.5912217097156
1.6 19.5872143191547
1.7 19.6015499757739
1.8 19.6220403987683
1.9 19.6468815325713
2 19.6864651547928
2.1 19.7273334640839
2.2 19.749063542732
2.3 19.7721909596248
2.4 19.73619855794
2.5 19.6568352306863
2.6 19.5822949480203
2.7 19.5213681492898
2.8 19.476258910304
2.9 19.5355551020404
3 19.6624649007422
3.1 19.8078709656765
3.2 19.92785101938
3.3 20.0347213072027
3.4 20.0618606895068
3.5 20.0419275007605
3.6 19.9986136309419
3.7 19.9504065529607
3.8 19.9043687116997
3.9 19.8997522451765
4 19.9372392468921
4.1 19.9924219961737
4.2 20.050868769618
4.3 20.1024206017755
4.4 20.1132598511985
4.5 20.1196921773452
4.6 20.1222504989224
4.7 20.1036336243278
4.8 20.095444190186
4.9 20.0609950066811
5 20.0365358492948
5.1 20.0153177072026
5.2 19.9987193172768
5.3 20.0041550160885
5.4 20.0244811472239
5.5 20.0532309780713
5.6 20.083206419401
5.7 20.1146713897216
5.8 20.1431979531982
5.9 20.1907436643829
6 20.2408816348981
6.1 20.2842964367388
6.2 20.3005748885273
6.3 20.3218128803697
6.4 20.3282693070683
6.5 20.3283536499443
6.6 20.3281485478863
6.7 20.3279832374161
6.8 20.3303931008715
6.9 20.3454049119839
7 20.3463565121556
7.1 20.3407601779881
7.2 20.3355428895543
7.3 20.3533330444283
7.4 20.366107767453
7.5 20.3953755351717
7.6 20.4326386852114
7.7 20.4495462593018
7.8 20.4745889582111
7.9 20.4917543223834
8 20.4826119755335
8.1 20.4825826265002
8.2 20.4912838323969
8.3 20.5097053116352
8.4 20.5043952813886
8.5 20.4917962914604
8.6 20.4844437511483
8.7 20.4871161058398
8.8 20.5220712061775
8.9 20.5334624711427
9 20.5384687949145
9.1 20.5677422081748
9.2 20.5944506965707
9.3 20.6253112262886
9.4 20.6616070097082
9.5 20.705667150998
9.6 20.7744710643799
};
\addlegendentry{Kalman Filter}
\addplot [color=red, line width= 1.0 pt]
table {%
0 20.1769414683517
0.1 20.1665957654071
0.2 20.1431463356251
0.3 20.1087506756272
0.4 20.0728319061697
0.5 20.0390760578732
0.6 20.0082824500756
0.7 19.9806907983224
0.8 19.9563878981525
0.9 19.9348721682812
1 19.9153438194472
1.1 19.8976794859497
1.2 19.8822745820258
1.3 19.8690037234332
1.4 19.8582066399404
1.5 19.8500453409715
1.6 19.8442172906001
1.7 19.840014733738
1.8 19.8370146616863
1.9 19.8347520189544
2 19.8328601840433
2.1 19.8314255841951
2.2 19.8308712833601
2.3 19.8315457622516
2.4 19.8353005928434
2.5 19.8430500247682
2.6 19.8546105434967
2.7 19.8696552429244
2.8 19.8871105808748
2.9 19.9039946118161
3 19.9188183597231
3.1 19.9312423783121
3.2 19.9416963504383
3.3 19.9506112283154
3.4 19.9597802963096
3.5 19.9699493348795
3.6 19.9814004700558
3.7 19.9940577514415
3.8 20.0076150939978
3.9 20.0209068619631
4 20.0330661506037
4.1 20.0439024737628
4.2 20.0534374461228
4.3 20.0620085025638
4.4 20.0704325704857
4.5 20.0786030105084
4.6 20.0867372092346
4.7 20.0953441791989
4.8 20.1044284572518
4.9 20.1147650273769
5 20.1261973155512
5.1 20.1388511563265
5.2 20.1526514792699
5.3 20.1671584691098
5.4 20.182111734513
5.5 20.1973397396286
5.6 20.2127537268371
5.7 20.2281424599491
5.8 20.243288008004
5.9 20.2575525651643
6 20.2708825273478
6.1 20.2834724446612
6.2 20.2957432990037
6.3 20.307491684691
6.4 20.319070487607
6.5 20.3305860772076
6.6 20.3420142497453
6.7 20.3533680293323
6.8 20.3645793683162
6.9 20.3755444829903
7 20.3867414796733
7.1 20.3982831096494
7.2 20.4100062426433
7.3 20.4213949633655
7.4 20.4325366874725
7.5 20.4429674368807
7.6 20.4526403566279
7.7 20.461947672211
7.8 20.470671962576
7.9 20.4790981254641
8 20.4876635765059
8.1 20.4959626904166
8.2 20.503805087589
8.3 20.5112336785294
8.4 20.518945615087
8.5 20.5270030966704
8.6 20.5351504287515
8.7 20.5430107988194
8.8 20.5500876071967
8.9 20.5570315978405
9 20.5636484915298
9.1 20.5692107719865
9.2 20.5736479738223
9.3 20.576645806034
9.4 20.5779977564454
9.5 20.577903590977
9.6 20.5778492215325
};
\addlegendentry{MHE}
\addplot [color=blue,dotted, line width= 1.0 pt]
table {%
0 21.9890475273132
0.1 21.9667530059814
0.2 21.9667410850525
0.3 21.9629192352295
0.4 19.6846008300781
0.5 19.5698523521423
0.6 19.6764135360718
0.7 19.6856641769409
0.8 19.560239315033
0.9 19.6514248847961
1 19.7638463973999
1.1 19.7469174861908
1.2 19.6196854114532
1.3 19.7295498847961
1.4 19.6237182617188
1.5 19.570837020874
1.6 19.596825838089
1.7 19.7215569019318
1.8 19.7084856033325
1.9 19.7584974765778
2 19.8480355739594
2.1 19.8388183116913
2.2 19.7386240959167
2.3 19.8540806770325
2.4 19.4312727451324
2.5 19.3165409564972
2.6 19.44011926651
2.7 19.4455409049988
2.8 19.4455409049988
2.9 20.0918340682983
3 20.2283501625061
3.1 20.2229189872742
3.2 20.1145172119141
3.3 20.2372717857361
3.4 19.8212218284607
3.5 19.8212206363678
3.6 19.8174679279327
3.7 19.8305010795593
3.8 19.8173916339874
3.9 20.0615572929382
4 20.1961398124695
4.1 20.1961398124695
4.2 20.2054762840271
4.3 20.2014493942261
4.4 19.9907231330872
4.5 20.1263284683228
4.6 20.1223349571228
4.7 19.9912059307098
4.8 20.1274752616882
4.9 19.8883128166199
5 20.026261806488
5.1 19.9891245365143
5.2 19.9887454509735
5.3 20.1102542877197
5.4 20.1402950286865
5.5 20.1638841629028
5.6 20.1690053939819
5.7 20.2056646347046
5.8 20.2106666564941
5.9 20.3867769241333
6 20.3790855407715
6.1 20.3711938858032
6.2 20.2520203590393
6.3 20.4057240486145
6.4 20.2937698364258
6.5 20.3125882148743
6.6 20.3363609313965
6.7 20.3383874893188
6.8 20.3570938110352
6.9 20.4427409172058
7 20.3140687942505
7.1 20.3233098983765
7.2 20.3473472595215
7.3 20.4998302459717
7.4 20.3879523277283
7.5 20.5347371101379
7.6 20.5569815635681
7.7 20.4218602180481
7.8 20.5718517303467
7.9 20.5164337158203
8 20.3836727142334
8.1 20.5348539352417
8.2 20.5626368522644
8.3 20.6072354316711
8.4 20.4169201850891
8.5 20.4519033432007
8.6 20.5059623718262
8.7 20.5504083633423
8.8 20.7394528388977
8.9 20.4826784133911
9 20.5402207374573
9.1 20.7388734817505
9.2 20.661039352417
9.3 20.7277250289917
9.4 20.7816314697266
9.5 20.8522725105286
9.6 21.0379409790039
};
\addlegendentry{Naive}
\end{axis}

\end{tikzpicture}}
        \caption{}
    \end{subfigure}
    \\
    \begin{subfigure}[b]{0.48\textwidth}
        \centering
        \resizebox{0.9\textwidth}{!}{
\begin{tikzpicture}

\definecolor{darkgray176}{RGB}{176,176,176}
\definecolor{darkorange25512714}{RGB}{255,127,14}
\definecolor{forestgreen4416044}{RGB}{44,160,44}
\definecolor{lightgray204}{RGB}{204,204,204}
\definecolor{steelblue31119180}{RGB}{31,119,180}

\begin{axis}[
legend cell align={left},
legend style={fill opacity=0.8, draw opacity=1, text opacity=1, draw=lightgray204, legend columns=3},
tick align=outside,
tick pos=left,
x grid style={darkgray176},
xlabel={time (sec)},
xmajorgrids,
xmin=0, xmax=10,
xtick style={color=black},
y grid style={darkgray176},
ylabel={Speed (m/s)},
ymajorgrids,
ymin=22, ymax=36,
ytick style={color=black},
ytick={22,24,26,28,30,32,34,36},
yticklabels={22,24,26,28,30,32,34,36}
]
\addplot [color=black,  dashed,line width= 1.0 pt]
table {
0 28.393775331453
0.1 29.1158072346335
0.2 29.268234137392
0.3 29.3348131303666
0.4 29.4288584240499
0.5 29.5147035998663
0.6 29.5939015502844
0.7 29.6623431160005
0.8 29.7035654179263
0.9 29.7499830582391
1 29.7899595651582
1.1 29.8293868563301
1.2 29.8628261983959
1.3 29.8881636378144
1.4 30.3859948149729
1.5 31.0571519365569
1.6 31.7368329942279
1.7 32.3269699971401
1.8 32.7913283230059
1.9 32.6897195570806
2 32.2933060576285
2.1 31.7913142631502
2.2 31.2931918339442
2.3 30.8669617083905
2.4 30.3381608042392
2.5 29.8187424470946
2.6 29.4598455399639
2.7 29.2365317665018
2.8 29.1193801664689
2.9 29.125757665996
3 29.2115432332307
3.1 29.3289770875597
3.2 29.4337000560482
3.3 29.5297912295299
3.4 29.6135423671254
3.5 29.6740046149703
3.6 29.7183279953856
3.7 29.7378128434706
3.8 29.7411087180139
3.9 29.8087858116061
4 29.908870826043
4.1 30.0142080882137
4.2 30.113871972396
4.3 30.1980072859145
4.4 30.2356309428686
4.5 30.2379531774491
4.6 30.2258770870189
4.7 30.2013832384302
4.8 30.185073871426
4.9 30.1648261400411
5 30.1486955816344
5.1 30.140511210124
5.2 30.1447560774788
5.3 30.1555894472031
5.4 30.1770043368093
5.5 30.1955148289416
5.6 30.205797574869
5.7 30.1655427447293
5.8 30.1007559018184
5.9 29.11803019846
6 27.7740541528038
6.1 27.3098048068125
6.2 27.3900581541862
6.3 27.7520583689889
6.4 28.3135751027785
6.5 28.9134202121763
6.6 29.4540543467898
6.7 29.7738290145675
6.8 29.5925248851854
6.9 29.1554353649375
7 28.6415260239491
7.1 28.5365101804165
7.2 28.6675684912279
7.3 28.9090788714129
7.4 29.1547217591
7.5 29.3677950088047
7.6 29.5343764642119
7.7 29.6523992672882
7.8 29.7247944583691
7.9 29.7628772696449
8 29.7742060376956
8.1 29.7703602225605
8.2 29.761753348719
8.3 29.7680721727859
8.4 29.7800247386767
8.5 29.7929646354078
8.6 29.7673035723618
8.7 29.7139660187697
8.8 29.651422783884
8.9 29.5985743820156
9 29.556125651419
9.1 29.5623746958898
9.2 29.5922586011593
9.3 29.5590665153481
9.4 29.4896520454249
9.5 29.4890658449397
9.6 29.5254100479387
};
\addlegendentry{Kalman Filter}
\addplot [color=red,  line width= 1.0 pt]
table {%
0 30.1255188896752
0.1 30.1300353407321
0.2 30.1406130264684
0.3 30.1572049359939
0.4 30.1786478595219
0.5 30.2046224845787
0.6 30.2349671435926
0.7 30.2698776887905
0.8 30.3095380935017
0.9 30.3535919870932
1 30.4017099288114
1.1 30.4532129716726
1.2 30.5064603487044
1.3 30.5562262697634
1.4 30.5877291609422
1.5 30.5960021956223
1.6 30.5795262051078
1.7 30.5391735405896
1.8 30.4793603932081
1.9 30.4134402466963
2 30.3463980635791
2.1 30.2801431961371
2.2 30.2159003361899
2.3 30.1557108720444
2.4 30.1052945995229
2.5 30.0658059914637
2.6 30.0350806954668
2.7 30.0120050140672
2.8 29.9953534316293
2.9 29.9829865342702
3 29.9736502627567
3.1 29.9668020501834
3.2 29.9624506875125
3.3 29.9602318601233
3.4 29.9598643068033
3.5 29.9613610608356
3.6 29.964560566391
3.7 29.9693343095868
3.8 29.9748584576463
3.9 29.9786528503995
4 29.9795381222284
4.1 29.9770123014577
4.2 29.9706903080455
4.3 29.9606444216789
4.4 29.9476093961343
4.5 29.9318668010739
4.6 29.9133073774225
4.7 29.891801093898
4.8 29.8667951938062
4.9 29.8380832572887
5 29.8051523581953
5.1 29.7675109553675
5.2 29.7244268817888
5.3 29.6753278058147
5.4 29.6195799815382
5.5 29.557550475959
5.6 29.4910108395233
5.7 29.4247700284703
5.8 29.368753703617
5.9 29.3479431559394
6 29.362347959718
6.1 29.3872630615863
6.2 29.4128147925349
6.3 29.4343202938591
6.4 29.4466763930011
6.5 29.4501309888688
6.6 29.4468911180372
6.7 29.4436016931068
6.8 29.4525701382096
6.9 29.476731471043
7 29.5133332468865
7.1 29.5514918485651
7.2 29.5872164468915
7.3 29.6193129656351
7.4 29.6475743296381
7.5 29.672615786462
7.6 29.69485799826
7.7 29.7142582222253
7.8 29.7309225465103
7.9 29.7452886192518
8 29.7573878162699
8.1 29.7676053003733
8.2 29.7760351143719
8.3 29.7826915699602
8.4 29.7876476262833
8.5 29.7913215607711
8.6 29.7948272781314
8.7 29.7981932795367
8.8 29.8014628598351
8.9 29.8046757352642
9 29.8073486314913
9.1 29.8078978756643
9.2 29.8066487170961
9.3 29.8060705905689
9.4 29.8064307699101
9.5 29.8063451462254
9.6 29.806292987689
};
\addlegendentry{MHE}
\addplot [color=blue, dotted,line width= 1.0 pt]
table {%
0 29.3274998664856
0.1 29.3835258483887
0.2 29.3680119514465
0.3 29.4310474395752
0.4 29.7576451301575
0.5 29.7484564781189
0.6 29.7926449775696
0.7 29.7981643676758
0.8 29.6980094909668
0.9 29.878785610199
1 29.857165813446
1.1 29.9179172515869
1.2 29.9164223670959
1.3 29.9150323867798
1.4 33.3889579772949
1.5 33.2858777046204
1.6 33.3375978469849
1.7 33.3361768722534
1.8 33.3684635162354
1.9 30.2332973480225
2 30.2248215675354
2.1 30.2245020866394
2.2 30.1672410964966
2.3 30.1745629310608
2.4 28.722186088562
2.5 28.6648797988892
2.6 29.2821002006531
2.7 29.2743802070618
2.8 29.2877626419067
2.9 29.6547651290894
3 29.7553157806396
3.1 29.7611093521118
3.2 29.6592974662781
3.3 29.7596192359924
3.4 29.8017287254333
3.5 29.7646999359131
3.6 29.8072504997253
3.7 29.727566242218
3.8 29.7275638580322
3.9 30.2727627754211
4 30.3301358222961
4.1 30.3437399864197
4.2 30.4007601737976
4.3 30.4137873649597
4.4 30.2204084396362
4.5 30.1773428916931
4.6 30.2267575263977
4.7 30.1699662208557
4.8 30.2833962440491
4.9 30.2099585533142
5 30.2534770965576
5.1 30.2581000328064
5.2 30.3284239768982
5.3 30.3348684310913
5.4 30.4528188705444
5.5 30.3926706314087
5.6 30.3166913986206
5.7 29.9616432189941
5.8 29.9082136154175
5.9 23.3457112312317
6 23.3594369888306
6.1 29.7974634170532
6.2 29.8414921760559
6.3 29.776623249054
6.4 30.8763122558594
6.5 30.5327796936035
6.6 30.5541181564331
6.7 30.0142312049866
6.8 27.2695922851562
6.9 27.3576164245605
7 27.4155187606812
7.1 30.045690536499
7.2 30.0785350799561
7.3 30.045530796051
7.4 30.0823950767517
7.5 29.9367904663086
7.6 29.8991060256958
7.7 29.9858689308167
7.8 30.0239253044128
7.9 29.9223399162292
8 30.0128126144409
8.1 29.9160695075989
8.2 29.9366140365601
8.3 29.9555039405823
8.4 29.9850034713745
8.5 29.9849247932434
8.6 29.7009897232056
8.7 29.8046278953552
8.8 29.8136830329895
8.9 29.7589874267578
9 29.7533893585205
9.1 30.2161741256714
9.2 30.2082896232605
9.3 29.5168232917786
9.4 29.5461106300354
9.5 30.0358963012695
9.6 30.0358867645264
};
\addlegendentry{Naive}
\end{axis}

\end{tikzpicture}}
        \caption{}
        \label{subfig:speed_plot_c}
    \end{subfigure}
    \begin{subfigure}[b]{0.48\textwidth}
        \centering
        \resizebox{0.9\textwidth}{!}{
\begin{tikzpicture}

\definecolor{darkgray176}{RGB}{176,176,176}
\definecolor{darkorange25512714}{RGB}{255,127,14}
\definecolor{forestgreen4416044}{RGB}{44,160,44}
\definecolor{lightgray204}{RGB}{204,204,204}
\definecolor{steelblue31119180}{RGB}{31,119,180}

\begin{axis}[
legend cell align={left},
legend style={fill opacity=0.8, draw opacity=1, text opacity=1, draw=lightgray204, legend columns=1},
tick align=outside,
tick pos=left,
x grid style={darkgray176},
xlabel={time (sec)},
xmajorgrids,
xmin=0, xmax=10,
xtick style={color=black},
y grid style={darkgray176},
ylabel={Speed (m/s)},
ymajorgrids,
ymin=24, ymax=42,
ytick style={color=black},
ytick={28,26,30,32,34,36,38,40,42},
yticklabels={28,26,30,32,34,36,38,40,42}
]
\addplot [color=black, dashed,line width= 1.0 pt]
table {
0 25.1404246935243
0.1 25.1618171276871
0.2 25.1735397974768
0.3 25.1819704303543
0.4 25.2113913998977
0.5 25.2368270336422
0.6 25.2584499448093
0.7 25.2749623964753
0.8 25.2895591243558
0.9 25.3074249017463
1 25.3253069237112
1.1 25.3409569152722
1.2 25.3557136976036
1.3 25.3685853030418
1.4 25.3831020942891
1.5 25.3965420455719
1.6 25.4090301795422
1.7 25.4208729842493
1.8 25.4326920340875
1.9 25.4374898512964
2 25.4401249047714
2.1 25.4423504049185
2.2 25.4445335337586
2.3 25.4477814991279
2.4 25.4473544255819
2.5 25.4469581038503
2.6 25.4466963161144
2.7 25.4480216006086
2.8 25.4499316952254
2.9 25.4599216172471
3 25.4944455706037
3.1 25.4799360173077
3.2 25.4848771351546
3.3 25.4970845710157
3.4 25.5016305097283
3.5 25.5041578455549
3.6 25.5060342124907
3.7 25.5083491147063
3.8 25.5116765561699
3.9 25.525164656265
4 25.5425222877116
4.1 25.5602155312745
4.2 25.577170596127
4.3 25.594065211944
4.4 25.6071602923286
4.5 25.6185927094057
4.6 25.6297689788641
4.7 25.6407601811297
4.8 26.4906382101855
4.9 26.7324644604391
5 26.7488454842928
5.1 26.6872809555486
5.2 26.602387961334
5.3 26.5156238095986
5.4 26.4188958653972
5.5 26.3244478365242
5.6 26.2376223852189
5.7 26.1582930216168
5.8 26.0881051698726
5.9 26.0258229768873
6 25.9700570654031
6.1 25.9209814437227
6.2 25.8772160734079
6.3 25.8388665778879
6.4 25.801665031243
6.5 25.768784367606
6.6 25.7400061041593
6.7 25.71306476079
6.8 25.6889830224129
6.9 25.6664912903627
7 25.6457543723682
7.1 25.6278540487097
7.2 25.6118672749065
7.3 25.5981069979973
7.4 25.5813431004286
7.5 25.5632143526161
7.6 25.5464684517282
7.7 25.5314043125626
7.8 25.518410297165
7.9 25.5139235398727
8 25.5132286507346
8.1 25.5125405819701
8.2 25.5157408298527
8.3 25.5194938254865
8.4 25.5295233337081
8.5 25.5437561840454
8.6 25.5596061195195
8.7 25.5755159931611
8.8 25.5911332822315
8.9 25.5195963898661
9 25.4222784772686
9.1 25.3221897631544
9.2 25.229910774202
9.3 25.1445504933588
9.4 25.1167264062886
9.5 25.108677333974
9.6 25.1091571178309
};
\addlegendentry{Kalman Filter}
\addplot [color=red,  line width= 1.0 pt]
table {%
0 25.6084679138508
0.1 25.6094041967731
0.2 25.6116773397279
0.3 25.6151576232195
0.4 25.6191299666102
0.5 25.6233245714569
0.6 25.6276862817905
0.7 25.6320713228622
0.8 25.6363406685305
0.9 25.6403256037506
1 25.6439655647448
1.1 25.6473329173021
1.2 25.650574468172
1.3 25.6538446748262
1.4 25.6572293398491
1.5 25.6607834292203
1.6 25.6644415784891
1.7 25.668447722137
1.8 25.6728446898406
1.9 25.678263738935
2 25.6848791326361
2.1 25.6927945602685
2.2 25.7020091708541
2.3 25.7125221763993
2.4 25.7245145306456
2.5 25.7379942907816
2.6 25.7530058325905
2.7 25.7695277748385
2.8 25.787319648385
2.9 25.8055820798725
3 25.823668413919
3.1 25.8438474136092
3.2 25.864990211607
3.3 25.8869482195731
3.4 25.9104924894358
3.5 25.9360196378432
3.6 25.9637800931396
3.7 25.9938485375083
3.8 26.0262379382994
3.9 26.0603851983994
4 26.0961475819794
4.1 26.1338535613151
4.2 26.1737910332307
4.3 26.2159540991204
4.4 26.2600829173214
4.5 26.3043772123665
4.6 26.3433990841752
4.7 26.3622151949471
4.8 26.3212477190565
4.9 26.2612495191405
5 26.1973877287625
5.1 26.1350893064664
5.2 26.0761551721348
5.3 26.0212284230323
5.4 25.9711034313251
5.5 25.9257816801888
5.6 25.8848541635923
5.7 25.8479117643181
5.8 25.8144419318816
5.9 25.7840584654671
6 25.7564333758452
6.1 25.7311995373523
6.2 25.7080456220988
6.3 25.6867018031693
6.4 25.6670257798495
6.5 25.6487028019002
6.6 25.6314861396275
6.7 25.6153048116259
6.8 25.5999713194262
6.9 25.5852958407206
7 25.5711265277959
7.1 25.5572843947664
7.2 25.5436785440755
7.3 25.5302836642256
7.4 25.517415136358
7.5 25.5051807959526
7.6 25.4934842864395
7.7 25.4821241906192
7.8 25.4707520169498
7.9 25.4586454550795
8 25.4454062007687
8.1 25.4307552937341
8.2 25.4141410799955
8.3 25.3951080637209
8.4 25.3728605997019
8.5 25.3468630129255
8.6 25.3169749491917
8.7 25.2837284634814
8.8 25.2492420101192
8.9 25.2197246179845
9 25.1972739280935
9.1 25.1823136206492
9.2 25.1743064540085
9.3 25.1718550438089
9.4 25.1712233945417
9.5 25.1711123410585
9.6 25.171048239701
};
\addlegendentry{MHE}
\addplot [color=blue,dotted, line width= 1.0 pt]
table {%
0 25.4089665412903
0.1 25.463445186615
0.2 25.4237604141235
0.3 25.4337024688721
0.4 25.6101965904236
0.5 25.6228756904602
0.6 25.5885767936707
0.7 25.6413888931274
0.8 25.6067109107971
0.9 25.6798052787781
1 25.6446027755737
1.1 25.6401991844177
1.2 25.6576442718506
1.3 25.6532144546509
1.4 25.6582927703857
1.5 25.6510782241821
1.6 25.7024717330933
1.7 25.6212735176086
1.8 25.6801414489746
1.9 25.5085277557373
2 25.5563545227051
2.1 25.5639410018921
2.2 25.5663657188416
2.3 25.5863308906555
2.4 25.5489444732666
2.5 25.5687952041626
2.6 25.5692887306213
2.7 25.5944299697876
2.8 25.5950736999512
2.9 25.7541871070862
3 26.1189293861389
3.1 25.2612662315369
3.2 25.8300280570984
3.3 25.8804535865784
3.4 25.691020488739
3.5 25.7054924964905
3.6 25.7075357437134
3.7 25.7194399833679
3.8 25.7037234306335
3.9 25.8904337882996
4 25.9402084350586
4.1 25.9086942672729
4.2 25.8812713623047
4.3 25.9302401542664
4.4 25.8720231056213
4.5 25.8782029151917
4.6 25.8938789367676
4.7 25.9030032157898
4.8 40.3767108917236
4.9 25.6689763069153
5 25.6654691696167
5.1 25.6838226318359
5.2 25.6833243370056
5.3 25.7017636299133
5.4 25.4711556434631
5.5 25.451557636261
5.6 25.4888701438904
5.7 25.4665899276733
5.8 25.5130362510681
5.9 25.5080008506775
6 25.5073094367981
6.1 25.526225566864
6.2 25.5256485939026
6.3 25.5411982536316
6.4 25.4953384399414
6.5 25.5380725860596
6.6 25.5474328994751
6.7 25.5121660232544
6.8 25.5217719078064
6.9 25.5319166183472
7 25.5288624763489
7.1 25.5451035499573
7.2 25.5424070358276
7.3 25.552179813385
7.4 25.4534959793091
7.5 25.4217791557312
7.6 25.4344129562378
7.7 25.4380536079407
7.8 25.4473519325256
7.9 25.5777764320374
8 25.5872344970703
8.1 25.5588173866272
8.2 25.6344175338745
8.3 25.6187677383423
8.4 25.7399988174438
8.5 25.7869172096252
8.6 25.8033180236816
8.7 25.810124874115
8.8 25.8201289176941
8.9 24.3453478813171
9 24.3497490882874
9.1 24.3567180633545
9.2 24.4049453735352
9.3 24.3842887878418
9.4 25.2332973480225
9.5 25.203104019165
9.6 25.2227163314819
};
\addlegendentry{Naive}
\end{axis}

\end{tikzpicture}}
        \caption{}
        \label{subfig:speed_plot_d}
    \end{subfigure}
    \caption{Estimation of speed for four different non-ego vehicles across various logs. The blue plot represents the \gls{mhe} estimates, whereas the orange plot denotes the naive speed estimate, obtained by dividing the distances and times in between the annotation intervals.}
    \label{fig:speed_plot}
\end{figure*}

A comparison between \gls{mhe}, a Kalman Filter~\cite{kalman1960new} based estimation, and a basic speed estimation approach is depicted in \Cref{fig:speed_plot}. Four non-ego vehicle tracks are sampled at random across various logs. For convenience, the tracks are chosen such that they are visible for the major duration of the sequence. 
The \gls{kf} estimate, denoted by dotted black, maintains the same state space and the state transition model as MHE. 
The basic, or naive speed estimate shown in blue, is obtained by simply dividing the distances and times between the annotation intervals. The \gls{mhe} estimate, shown in red, follows a smooth trajectory due to the stage cost in \cref{eq:mhe2} being constrained by the kinematic model. On the other hand, the naive estimate follows an irregular speed curve, which is due to the human annotator labelling differing and inconsistent views of the object at different time instances. \gls{mhe} also helps in removing outliers in naive speed estimate in extreme cases (\cref{subfig:speed_plot_c,subfig:speed_plot_d}).
The recusrive \gls{kf} estimate is observed to be less smooth compared to MHE.
Although a detailed comparison is challenging due to lack of precise ground truth, the smoother motion produced by MHE estimates makes it a more appropriate method.

\begin{figure*}[htbp!] 
\centering
\includegraphics[width=1.\linewidth]{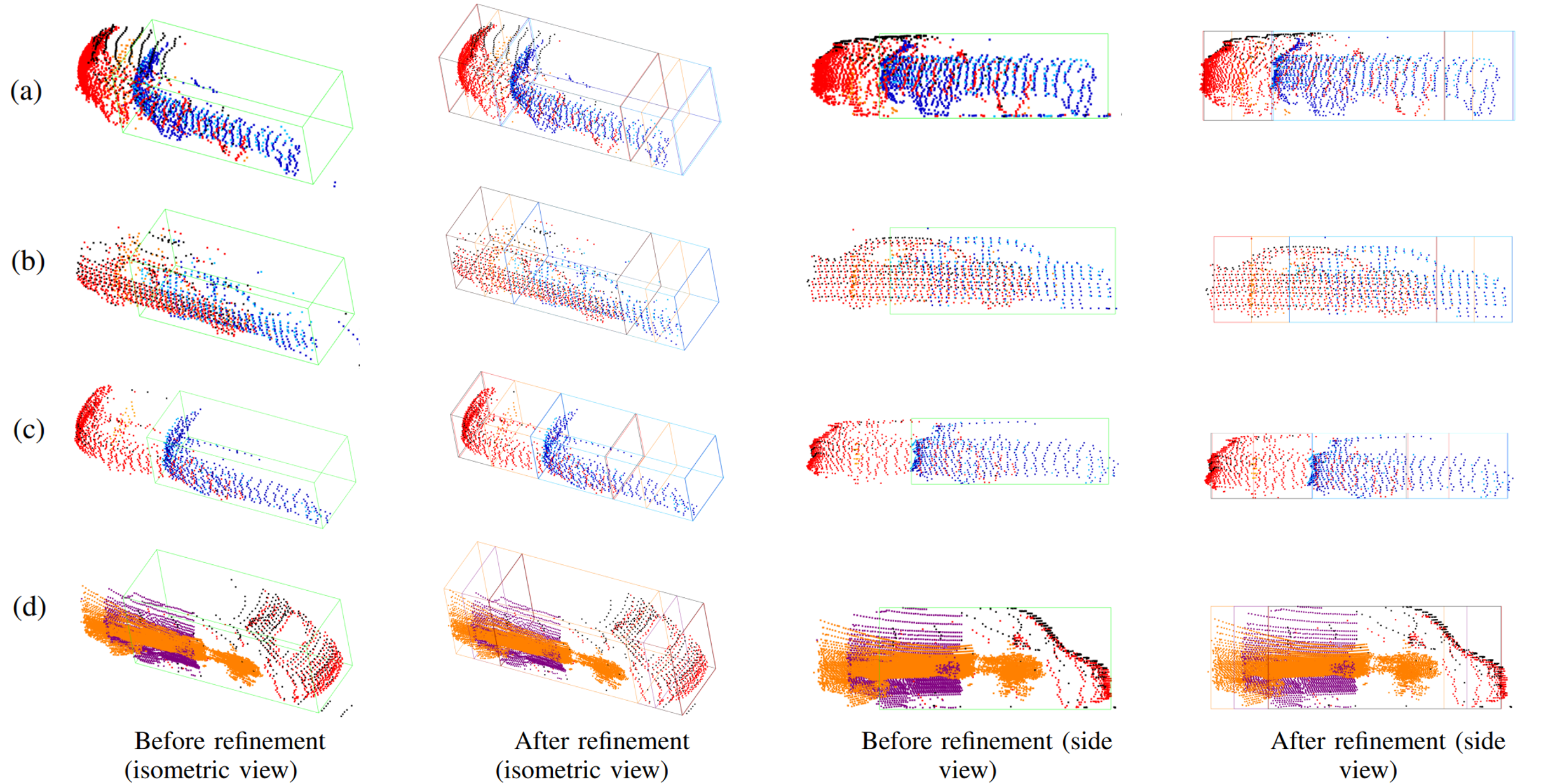}
\caption{Comparison of annotations before and after refinement proposed in \cref{subsec:annotation_refinement}. Rows 1-4 represent objects whose speed plots are given in Fig. 5 (a)-(d) respectively. The green box represents human annotated bounding box. 
 The point colors distinguish points captured by different \gls{lidar} sensors. Following refinement, the pseudo bounding boxes are color-coded based on the 
 point clusters that they are aligned with. The pseudo bounding boxes successfully capture missing views of the object.
 } 
 \label{fig:annotation_refinement}
\end{figure*}

The results of annotation refinement for the selected vehicles, as mentioned in the previous paragraph, are depicted in \Cref{fig:annotation_refinement}. The human-annotated box, colored in green, clearly misses various views, as indicated by the presence of red and black points in rows one to three and orange and violet points in row four. 
These discrepancies are effectively addressed by the refined annotations or pseudo bounding boxes, which are color-coded based on the point clusters they encompass, allowing them to accurately represent these perspectives. 
Of particular interest, rows 1-3 depict partially observed vehicles. The process of identifying high-density regions plays a pivotal role in classifying the rear side of the vehicle, facilitating the precise fitting of bounding boxes on speed-compensated clusters. This step was observed to be crucial for obtaining accurate pseudo bounding boxes.

\section{Conclusions and Future Work} 
\label{sec:conclusions}

This article has presented a solution to the data annotation challenges associated with heavy vehicles equipped with multiple active sensors. We have utilized \gls{mhe} estimators to estimate the speed of non-ego objects and rectify bounding boxes. The effectiveness of the proposed solution is demonstrated through an evaluation using real-life data gathered and annotated by Scania. Looking ahead, there are several avenues for further research and development. One such direction involves tailoring the modeling approach based on the specific class of objects, such as bicycle models, articulated vehicles, pedestrians, and so on. This customization can potentially enhance the accuracy of \gls{mhe} speed estimation. 
Moreover, the current framework operates on 10~\unit{s} frames. Future research can focus on 
extending its application to longer time sequences without annotations, providing a cost-effective means to expand annotated datasets. 
A prior step for this would involve using the refined annotations to train \gls{dnn} algorithms for object detection, tracking etc.
In summary, the proposed solution represents a significant step forward in addressing data annotation challenges in heavy vehicle sensor systems. With ongoing research and innovation, we can anticipate further advancements in this field, contributing to the evolution of safer and more efficient transportation technologies and addressing a common bottleneck in developing machine learning models for autonomous vehicles and other applications.




\bibliographystyle{IEEEtran}
\bibliography{mybib}

\end{document}